\documentclass[letterpaper,10pt,conference]{style/ieeeconf}
\IEEEoverridecommandlockouts{}
\usepackage[T1]{fontenc}
\usepackage[dvipsnames]{xcolor}
\usepackage{graphicx,amsmath,amssymb,mathtools}
\usepackage[font=small]{caption}
\usepackage{siunitx} %
\usepackage{subcaption,booktabs,acro,multirow}
\usepackage{tikz,pgfplots,pgfplotstable}
\pgfplotsset{compat=1.18}
\usetikzlibrary{positioning,calc,backgrounds,arrows.meta}
\makeatletter
\let\NAT@parse\undefined
\makeatother
\usepackage[hidelinks]{hyperref}
\usepackage[capitalise,nameinlink]{cleveref}
\DeclareAcronym{ate}{short=ATE, long=Absolute Trajectory Error}
\DeclareAcronym{rte}{short=RTE, long=Relative Trajectory Error}
\DeclareAcronym{wrte}{short=RTE\(_j\), long=windowed Relative Trajectory Error}
\DeclareAcronym{icp}{short=ICP, long=Iterative Closest Point}
\DeclareAcronym{imu}{short=IMU, long=Inertial Measurement Unit}
\DeclareAcronym{lidar}{short=LiDAR, long=Light Detection and Ranging}
\DeclareAcronym{lio}{short=LIO, long=\ac{lidar}-Inertial Odometry}
\DeclareAcronym{lo}{short=LO, long=\ac{lidar} Odometry}
\DeclareAcronym{slam}{short=SLAM, long=Simultaneous Localization and Mapping}
\DeclareAcronym{atv}{short=ATV, long=All-Terrain Vehicle}
\DeclareAcronym{pca}{short=PCA, long=Principal Component Analysis}
\DeclareAcronym{sota}{short=SOTA, long=state-of-the-art}

\DeclareAcronym{ours}{short=xFORM, long=Accelerated \textbf{F}ixed-Lag \textbf{O}dometry with \textbf{R}eparative \textbf{M}apping}
\DeclareAcronym{ours_plain}{short=xFORM, long=Accelerated Fixed-Lag Odometry with Reparative Mapping}

\DeclareAcronym{n20}{short=N20, long=Newer Stereo-Cam}
\DeclareAcronym{n21}{short=N21, long=Newer Multi-Cam}
\DeclareAcronym{h22}{short=H22, long=Hilti 2022}
\DeclareAcronym{os}{short=OS, long=Oxford Spires}
\DeclareAcronym{mc}{short=MC, long=Multi-Campus}
\DeclareAcronym{cu}{short=CU, long=CU-Multi}
\DeclareAcronym{bg}{short=BG, long=Botanic Gardens}

\definecolor{stExtract}{HTML}{0173b2}
\definecolor{stMatch}{HTML}{029e73}
\definecolor{stSemi}{HTML}{de8f05}
\definecolor{stFull}{HTML}{9c4a06}
\definecolor{stOther}{HTML}{949494}
\newcommand{\CpuSpeedupRange}{1.1--1.4}
\newcommand{\GpuSpeedupRange}{2.0--4.0}

\newcommand{\AgreementPosition}{1.2\times10^{-8}}
\newcommand{\AgreementRotation}{3.4\times10^{-11}}
\newcommand{\TraceComparisonCount}{60}
\newcommand{\BenchmarkRunCount}{90}

\newcommand{\MaxActivePoses}{61}

\newcommand{\StairsFormMean}{132.9}

\newcommand{\StairsCpuMean}{107.3}

\newcommand{\StairsGpuMean}{32.9}

\newcommand{\WorkloadFeatureRange}{5,000--18,800}
\newcommand{\WorkloadMatchRange}{53,000--268,000}
\newcommand{\WorkloadPoseRange}{22--44}
\newcommand{\WorkloadPairRange}{230--950}
\newcommand{\WorkloadRematchRange}{10--21}
\newcommand{\WorkloadLmRange}{17--30}
\newcommand{\OverallCpuSpeedup}{1.2}
\newcommand{\OverallGpuSpeedup}{3.0}

\newcommand{\AblationStairsFormExtract}{19.3}

\newcommand{\AblationStairsCpuMean}{107.7}
\newcommand{\AblationStairsCpuExtract}{17.9}

\newcommand{\AblationStairsCpuMatch}{77.2}

\newcommand{\AblationStairsGpuMatchingMean}{43.3}

\newcommand{\AblationStairsGpuMean}{32.5}

\newcommand{\AblationOptimizerSemiForm}{19.54}
\newcommand{\AblationOptimizerSemiCompressed}{10.98}
\newcommand{\AblationOptimizerSemiSpeedup}{1.78}
\newcommand{\AblationOptimizerFullForm}{13.00}
\newcommand{\AblationOptimizerFullCompressed}{0.91}
\newcommand{\AblationOptimizerFullSpeedup}{14.21}
\newcommand{\AblationOptimizerCombinedForm}{32.54}
\newcommand{\AblationOptimizerCombinedCompressed}{11.89}
\newcommand{\AblationOptimizerCombinedSpeedup}{2.74}
\newcommand{\AblationOptimizerPipelineForm}{134.78}
\newcommand{\AblationOptimizerPipelineCompressed}{110.78}
\newcommand{\AblationOptimizerPipelineSpeedup}{1.22}

\title{xFORM\@: Accelerating Fixed-Lag LiDAR Odometry\\with Exact Factor Compression}
\author{Ruogu Li, Easton R. Potokar, and Michael Kaess%
\thanks{Authors are with the Robotics Institute, Carnegie Mellon University,
Pittsburgh, PA, USA\@. {\tt\footnotesize \{ruoguli,potokar,kaess\}@cmu.edu}.}}
\begin{document}
\maketitle
\begin{abstract}
    A increasingly popular sensor, \acf{lidar} have been successfully used for state estimation due to their precision, long range capabilities and high sensor rate.
    Unfortunately, \ac{lo} systems incur substantial computational cost when processing large point clouds, particularly those that jointly refine a window of scan poses.
    While our prior \ac{lo} work FORM is real-time, the costs of feature extraction, correspondence search, and the evaluation and linearization of residuals during pose optimization can introduce significant latency. This latency makes downstream tasks such as control and path planning difficult.
    To overcome this, we present xFORM, which accelerates the three stages of FORM\@.
    xFORM uses factor compression to reduce an optimization problem involving hundreds of thousands of feature matches to 34 scalar residuals per scan pair, while preserving the original objective and Gauss-Newton system in exact arithmetic.
    We integrate this optimizer into two variants: xFORM-CPU, which additionally parallelizes feature extraction on the CPU, and xFORM-GPU, which accelerates feature processing, correspondence search, and factor compression on the GPU\@.
    Across fifteen complete sequences from three datasets, xFORM-CPU and xFORM-GPU achieve overall speedups of \CpuSpeedupRange{}$\times$ and \GpuSpeedupRange{}$\times$ over FORM, respectively, with no numerically significant differences to the original implementation.
\end{abstract}
\section{Introduction}\label{sec:intro}

\acf{lidar} sensors have become the de facto sensor for many field robotic state estimation tasks due to their precision, reliability, and high speed scanning of environments. They have successfully been utilized in applications such as aerial vehicles, quadrapeds, and ground vehicles, with major improvements in accuracy and reliability over prior sensor modalities~\cite{leeLiDAROdometrySurvey2024}.

Most current \ac{sota} \ac{lo} methods unfortunately have properties that often invalidate their usage as input to the underlying vehicle controller. To be able to properly used as a control input, state estimates must be high rate, accurate, and produce smooth trajectories. We note that our prior work, FORM~\cite{potokarFORMFixedLagOdometry2025}, produces \ac{sota} accurate and smooth trajectories at real-time speeds, but often returns results at near \SI{100}{\milli\second} latency.

In feature-based LiDAR odometry, processing a scan involves three main stages: \emph{feature extraction}, \emph{correspondence search}, and \emph{pose optimization}.
Feature extraction selects points that represent useful geometric structures and computes properties such as surface normals.
Correspondence search finds matching features in previous scans or a local map.
Pose optimization uses these matches to estimate the sensor poses by minimizing geometric alignment errors, called residuals.
Across our evaluated sequences, FORM extracts \WorkloadFeatureRange{} feature points per scan on average, resulting in a final optimization containing \WorkloadMatchRange{} matches connecting \WorkloadPoseRange{} poses on average.
These numbers create substantial computational work in all three stages.

\begin{figure*}[t]
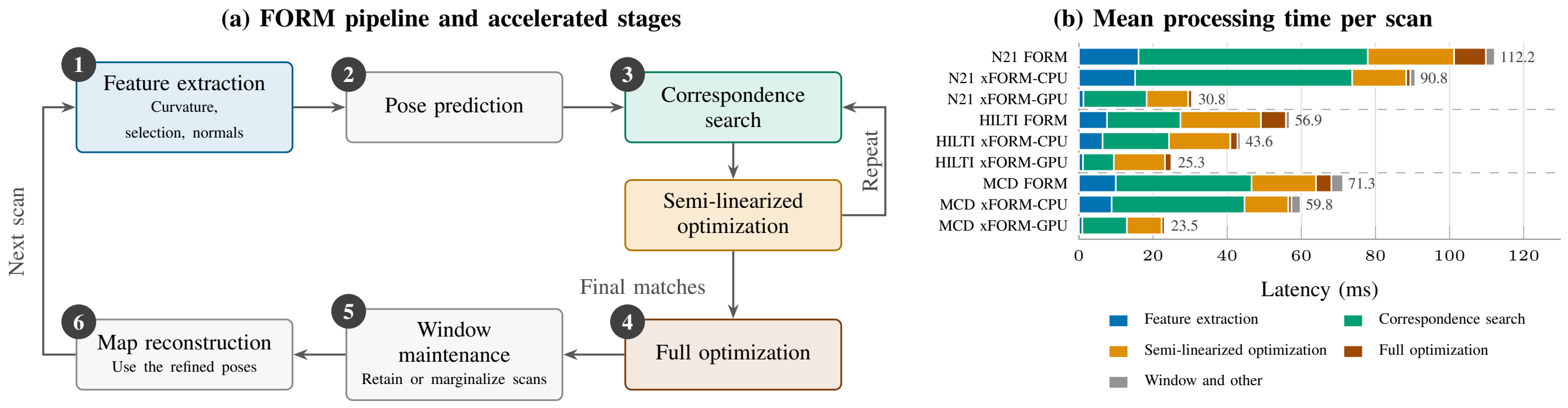

  \centering
  \begin{minipage}[t]{0.575\textwidth}\vspace{0pt}\centering
    {\footnotesize\textbf{(a) FORM pipeline and accelerated stages}}\\[5pt]
    \resizebox{\linewidth}{!}{\begin{tikzpicture}[
    x=1cm,y=1cm,font=\scriptsize,
    stage/.style={draw=black!45,rounded corners=2pt,line width=.55pt,
      align=center,inner sep=4pt,minimum height=.78cm,text width=2.12cm,fill=black!3},
    extract/.style={stage,fill=stExtract!12,draw=stExtract!80!black},
    match/.style={stage,fill=stMatch!13,draw=stMatch!75!black},
    semi/.style={stage,fill=stSemi!18,draw=stSemi!80!black},
    full/.style={stage,fill=stFull!12,draw=stFull!80!black},
    num/.style={circle,fill=black!75,text=white,font=\scriptsize\bfseries,
      inner sep=0pt,minimum size=11pt},
    arr/.style={-{Stealth[length=1.6mm]},line width=.65pt,draw=black!65},
    note/.style={font=\scriptsize,text=black!70,align=center}]

  \node[extract] (extract) at (1.65,2.85) {Feature extraction\\[-1pt]
    {\tiny Curvature, selection, normals}};
  \node[stage] (predict) at (4.65,2.85) {Pose prediction};
  \node[match] (match) at (7.75,2.85) {Correspondence search};
  \node[semi] (semi) at (7.75,1.65) {Semi-linearized\\optimization};

  \node[full] (full) at (7.75,.10) {Full optimization};
  \node[stage] (window) at (4.65,.10) {Window maintenance\\[-1pt]
    {\tiny Retain or marginalize scans}};
  \node[stage] (map) at (1.65,.10) {Map reconstruction\\[-1pt]
    {\tiny Use the refined poses}};

  \draw[arr] (extract) -- (predict);
  \draw[arr] (predict) -- (match);
  \draw[arr] (match) -- (semi);
  \draw[arr] (semi.east) -- ++(.48,0) |- (match.east);
  \node[note,rotate=90,anchor=south] at (9.53,2.25) {Repeat};
  \draw[arr] (semi) -- (full);
  \node[note,anchor=east] at (7.57,.86) {Final matches};
  \draw[arr] (full) -- (window);
  \draw[arr] (window) -- (map);
  \draw[arr] (map.west) -- (.08,.10) -- (.08,2.85) -- (extract.west);
  \node[note,rotate=90,anchor=south] at (-.01,1.47) {Next scan};

  \node[num] at ([xshift=1pt,yshift=-1pt]extract.north west) {1};
  \node[num] at ([xshift=1pt,yshift=-1pt]predict.north west) {2};
  \node[num] at ([xshift=1pt,yshift=-1pt]match.north west) {3};
  \node[num] at ([xshift=1pt,yshift=-1pt]full.north west) {4};
  \node[num] at ([xshift=1pt,yshift=-1pt]window.north west) {5};
  \node[num] at ([xshift=1pt,yshift=-1pt]map.north west) {6};

  \node[note,anchor=north] at (4.65,-.54)
    {Colored stages highlight the focus of xFORM.};
\end{tikzpicture}}
  \end{minipage}\hfill
  \begin{minipage}[t]{0.405\textwidth}\vspace{0pt}\centering
    {\footnotesize\textbf{(b) Mean processing time per scan}}\\[3pt]
    \pgfplotstableread{graphics/stage_breakdown.dat}\stagetable
\input{graphics/stage_axis.tex}
\begin{tikzpicture}[font=\scriptsize,
  totlabel/.style={anchor=west,font=\tiny,text=black!75,xshift=1.5pt,inner sep=0pt}]
\begin{axis}[
  xbar stacked, width=\dimexpr\linewidth-5pt\relax, height=3.8cm,
  bar width=5.6pt, xmin=0, xmax=\StageAxisMaximum,
  xlabel={Latency (ms)}, xlabel near ticks, xlabel style={font=\scriptsize},
  ytick=data, yticklabels from table={\stagetable}{label},
  y dir=reverse, ytick style={draw=none}, axis y line*=left, axis x line*=bottom,
  axis line style={black!40}, xtick style={black!40},
  enlarge y limits={abs=0.65}, tick label style={font=\tiny},
  xmajorgrids, grid style={black!12},
  legend style={at={(.5,-.31)},anchor=north,font=\tiny,draw=none,
    legend columns=2, column sep=3pt, row sep=1pt},
  legend cell align=left, legend image code/.code={\fill[#1] (0cm,-.07cm) rectangle (.22cm,.07cm);},
  every axis plot/.append style={draw=white,line width=.5pt},
  extra y ticks={2.5,5.5}, extra y tick labels={}, extra y tick style={grid=major,grid style={black!30,dashed}},
  clip=false]
\addplot[fill=stExtract] table[x=extract, y expr=\coordindex] {\stagetable}; \addlegendentry{Feature extraction}
\addplot[fill=stMatch]   table[x=match,   y expr=\coordindex] {\stagetable}; \addlegendentry{Correspondence search}
\addplot[fill=stSemi]    table[x=semi,    y expr=\coordindex] {\stagetable}; \addlegendentry{Semi-linearized optimization}
\addplot[fill=stFull]    table[x=full,    y expr=\coordindex] {\stagetable}; \addlegendentry{Full optimization}
\addplot[fill=stOther]   table[x=other,   y expr=\coordindex] {\stagetable}; \addlegendentry{Window and other}
\input{graphics/stage_totals.tex}
\end{axis}
\end{tikzpicture}
  \end{minipage}
  \vspace{-0.2em}
  \caption{(a) FORM's six per-scan steps. Step 3 alternates correspondence search and semi-linearized optimization until convergence or the iteration limit. Full optimization then refines all poses with the final matches fixed. Steps 5 and 6 maintain the pose window and reconstruct the map for the next scan. xFORM accelerates the colored stages through factor compression and CPU/GPU parallelism. (b) Mean latency on three datasets, equally weighted across five full sequences per dataset, divided by processing stage; N21 and MCD use two runs per configuration, and Hilti uses one. N21 denotes Newer College 2021 and MCD denotes Multi-Campus. Correspondence search includes map preparation. Using factor summaries in the optimizer reduces optimization time, while GPU processing accelerates extraction and matching.}
  \label{fig:overview}
  \vspace{-1.5em}
\end{figure*}

We present xFORM, which accelerates FORM through exact factor compression and optimized CPU and GPU implementation.
For each scan pair, xFORM compresses the fixed match data into small matrices called \emph{factor summaries}.
Evaluating these summaries at the current poses produces 34 scalar residuals per scan pair, regardless of the number of matches, while preserving the original nonlinear least-squares objective in exact arithmetic.
This reduces the work required to evaluate and linearize the residuals during optimization.
Summaries remain valid as the poses change and need rebuilding only when their matches change.

Beyond pose optimization, xFORM accelerates feature extraction and correspondence search through CPU parallelism and GPU processing.
\textbf{xFORM-CPU} parallelizes planar-feature selection across scanlines and evaluates and linearizes factor summaries in parallel across scan pairs.
\textbf{xFORM-GPU} extends this implementation with GPU feature processing and correspondence search, constructing factor summaries directly within the GPU matching pipeline.

These changes give xFORM-CPU and xFORM-GPU end-to-end speedups of \CpuSpeedupRange{}$\times$ and \GpuSpeedupRange{}$\times$, respectively, over FORM across the evaluated sequences.
For example, on Newer College Stairs, xFORM-CPU reduces mean processing time from \StairsFormMean{} to \StairsCpuMean{}\,ms per scan, and xFORM-GPU reduces it from \StairsFormMean{} to \StairsGpuMean{}\,ms while maintaining numerically identical results.

\section{Related Work}\label{sec:related}
As \acp{lidar} have become a standard sensor for robot state estimation, a plethora of \ac{lo} work has been done to find the most accurate, robust, and real-time method to accurately perform odometry.
LOAM~\cite{zhangLOAMLidarOdometry2014}, LeGO-LOAM~\cite{shanLeGOLOAMLightweightGroundOptimized2018}, KISS-ICP~\cite{vizzoKISSICPDefensePointtoPoint2023}, and GenZ-ICP~\cite{leeGenZICPGeneralizableDegeneracyRobust2025} register each scan against a submap and estimate one pose at a time, often resulting in jittery trajectories. In contrast, Liang et al.~\cite{liangHierarchicalEstimationBasedLiDAR2023} and FORM~\cite{potokarFORMFixedLagOdometry2025} keep a window of poses and refine them jointly, resulting in smoother trajectories, but often at significant latency. Other advanced techniques, such as continuous-time estimation~\cite{dellenbachCTICPRealtimeElastic2022,karimiLoLaSLAMLowLatencyLiDAR2021} or novel kd-tree structures~\cite{ferrariMADICPItAll2024} often introduce significant parameter tuning and latency as well.

Other work has been focused on solving this by summarizing point cloud matching residuals to increase computational speed.
Koide et al.~\cite{koideTightlyCoupledRange2025} selects a weighted subset of points that reproduces the quadratic registration model at one linearization pose. BALM and its successor~\cite{liuBALMBundleAdjustment2021,liuEfficientConsistentBundle2023} aggregate the points of a plane into fixed-size point-cluster statistics so that plane-adjustment cost and derivatives no longer depend on the point count. Our summary serves the same purpose for FORM's pairwise point-to-point and point-to-plane factors, keeps both endpoint poses as variables, and is exact at every pose rather than only at the pose where the subset was chosen.

Limited work has been done on utilizing GPU acceleration to reduce latency for \ac{lo}.
GLIM~\cite{koideGLIM3DRangeinertial2024} evaluates dense scan-matching factors on the GPU\@.
In contrast, xFORM solves the optimization on the CPU through factor compression and uses the GPU for feature extraction, correspondence search, and summary construction.

\section{Background: FORM}\label{sec:form}

We summarize FORM's pipeline and computational bottlenecks below, with full details in the original paper~\cite{potokarFORMFixedLagOdometry2025}.
FORM jointly estimates the poses of the $N_{\mathrm{recent}}$ most recent scans and a set of older scans, called \emph{keyscans}, that still have enough matches with recent scans to remain in the optimization.
Each pose maps a scan's points from that scan's local coordinate frame into a shared world frame.

\subsection{Per-Scan Pipeline}
Each incoming scan is processed in six steps (\cref{fig:overview}a).
\begin{enumerate}\setlength{\itemsep}{1pt}
    \item \emph{Feature extraction.} FORM selects two types of features from each scan: point features, represented by their 3D positions, and planar features, represented by their 3D positions and surface normals.
          To select these features, FORM divides each LiDAR scanline into angular sectors.
          Within each sector, it selects low-curvature points, which lie on locally smooth surfaces, as planar features and samples point features from the remaining valid points.
          Both selections enforce a minimum spacing between selected points.
          For each planar feature, FORM estimates a surface normal using neighboring points from adjacent scanlines.
    \item \emph{Pose prediction.} A constant-velocity model initializes the new scan's pose.
    \item \emph{Matching and semi-linearized optimization.} Each point or planar feature in the new scan is matched to the nearest feature of the same type in the local map, provided their distance is below a threshold.
          The map records each feature's source scan, so each match constrains the new pose and one earlier pose.
          Levenberg-Marquardt (LM) jointly refines all poses using registration \emph{factors}, which group the point-to-point and point-to-plane residuals between each pair of scans (\cref{sec:form-objective}).
          Factors between older scans retain a fixed linearization at the previous scan's optimized poses, while factors involving the new scan are relinearized at each LM iteration.
          Matching and optimization repeat until the new pose changes by less than a threshold or $N_{\mathrm{icp}}$ iterations (30 by default) are reached.
    \item \emph{Full optimization.} With the final matches fixed, LM minimizes the full nonlinear objective in \cref{eq:objective}, evaluating and relinearizing all registration factors at each iteration.
          The preceding semi-linearized stage saves computation while correspondences are changing.
          This final stage allows the older factors' linearizations to change as well, before the map is rebuilt.
    \item \emph{Window maintenance.} The oldest recent scan becomes a keyscan if it still matches enough recent scans.
          Otherwise, it is removed from the window and its information is retained as a prior on the remaining poses, a process called marginalization.
          Keyscans that no longer receive matches are also marginalized.
    \item \emph{Map reconstruction.} The retained scans' features are transformed into the world frame using the optimized poses to construct the map for the next scan.
\end{enumerate}

\subsection{Registration Objective}\label{sec:form-objective}
The matches from step 3 define the registration factors optimized in steps 3 and 4.
Let $X_i=(R_i,t_i)\in\mathrm{SE}(3)$ denote the rotation $R_i$ and translation $t_i$ of scan $i$ in the world frame.
A match consists of a query point $p_{j,a}$ in scan $j$'s frame and a target point $p_{i,a}$ in scan $i$'s frame.
For a planar target, its unit normal $n_{i,a}$ is also expressed in scan $i$'s frame.
The point-to-point and point-to-plane residuals are
\begin{align}
    r^{\mathrm{pt}}_{ij,a}(X_i,X_j) & =R_j p_{j,a} +t_j- R_i p_{i,a}-t_i\in\mathbb R^3,\label{eq:point}                \\
    r^{\mathrm{pl}}_{ij,a}(X_i,X_j) & =(R_i n_{i,a})^\top r^{\mathrm{pt}}_{ij,a}(X_i,X_j)\in\mathbb R.\label{eq:plane}
\end{align}
The \emph{factor} for a \emph{scan pair} $(i,j)$ stacks its $m_{\mathrm{pt}}$ point and $m_{\mathrm{pl}}$ plane residuals as
\begin{equation}
    \begin{aligned}
        r_{ij}(X_i,X_j) & =\operatorname{col}\!\left(
        \{r^{\mathrm{pt}}_{ij,a}\}_{a=1}^{m_{\mathrm{pt}}},
        \{r^{\mathrm{pl}}_{ij,a}\}_{a=1}^{m_{\mathrm{pl}}}\right)          \\
                        & \in\mathbb R^{3m_{\mathrm{pt}}+m_{\mathrm{pl}}},
    \end{aligned}\label{eq:stackedresidual}
\end{equation}
where $\operatorname{col}$ denotes vertical concatenation.
With isotropic noise scale $\sigma_{ij}$ per factor, FORM uses LM to solve the nonlinear least-squares problem
\begin{equation}
    \begin{aligned}
        X^\star & =\arg\min_X E(X),                                \\
        E(X)    & =E_{\mathrm{prior}}(X)+\sum_{(i,j)\in\mathcal E}
        \frac{\|r_{ij}(X_i,X_j)\|^2}{2\sigma_{ij}^2}.
    \end{aligned}\label{eq:objective}
\end{equation}
Here $X$ denotes all poses in the window, $\mathcal E$ is the set of scan pairs with matches, and $E_{\mathrm{prior}}$ contains pose priors and the information retained when earlier poses are marginalized.
For a fixed set of matches, the scan-frame points and normals remain constant while the poses are optimized.

\subsection{Computational Cost}
Feature extraction, matching, and optimization account for most of FORM's runtime (\cref{fig:overview}b).
Although optimization adjusts only a small window of scan poses (\WorkloadPoseRange{} poses on average across individual sequences in our evaluation), these poses are constrained by tens or hundreds of thousands of feature matches.
At each LM iteration, FORM evaluates alignment errors and computes their derivatives with respect to the poses to construct a linear system for the pose update.
It also evaluates the errors at proposed poses to decide whether to accept the update.
These computations repeatedly use the same matched points and surface normals, which remain fixed until correspondence search runs again.
Our factor summaries combine this fixed information into a compact representation, reducing the work required for subsequent error evaluations and derivative computations.

\begin{table}[t]
\centering
\caption{FORM's average per-scan workload, equally weighted across five full sequences per dataset and including warmup scans. Poses: active window (mean / maximum). Matches: final full optimization. LM: iterations per scan.}
\label{tab:workload}
\footnotesize
\setlength{\tabcolsep}{2.6pt}
\begin{tabular}{lrrrrrr}
\toprule
Dataset & Scanlines & Features & Poses & Matches & Rematches & LM \\
\midrule
N21 & 128 & 15,799 & 30 / 53 & 187,270 & 16.4 & 23.5 \\
Hilti & 32 & 5,842 & 35 / 61 & 98,516 & 11.3 & 19.9 \\
MCD & 64/128 & 9,333 & 28 / 52 & 94,326 & 16.9 & 23.2 \\
\bottomrule
\end{tabular}
\vspace{-1em}
\end{table}

\Cref{tab:workload} reports the workload averaged over five complete sequences from each of three datasets: Hilti (32 scanlines), Multi-Campus (MCD, 64/128 scanlines), and Newer College (128 scanlines).
Across individual sequences, the window contains \WorkloadPoseRange{} poses on average and at most \MaxActivePoses{}, but its factors contain \WorkloadMatchRange{} matches across \WorkloadPairRange{} scan pairs on average.
Processing one scan requires \WorkloadRematchRange{} rematches and \WorkloadLmRange{} LM iterations on average.

\section{Exact Factor Compression}\label{sec:summaries}

For a scan pair $(i,j)$, let $m_{\mathrm{pt}}$ and $m_{\mathrm{pl}}$ be the numbers of point and plane matches.
The registration residual vector $r_{ij}(X_i,X_j)$ in \cref{eq:stackedresidual} has $3m_{\mathrm{pt}}+m_{\mathrm{pl}}$ entries.
Our goal is to replace it with a vector $\tilde r_{ij}(X_i,X_j)$ containing 34 scalar residuals while preserving the quantities used by LM\@.
To specify these quantities, let $\delta_i,\delta_j\in\mathbb R^6$ be local pose increments, each comprising three rotation and three translation components.
Let $J_{ij}$ and $\tilde J_{ij}$ be the Jacobians of the original and compressed residual vectors with respect to the same stacked increment $\delta=(\delta_i^\top,\delta_j^\top)^\top\in\mathbb R^{12}$.
For fixed matches, we require the following identities at every pose pair $(X_i,X_j)$ in exact arithmetic:
\begin{equation}
  \begin{aligned}
    \tilde r_{ij}^{\top}\tilde r_{ij} & =r_{ij}^{\top}r_{ij}, \\
    \tilde J_{ij}^{\top}\tilde r_{ij} & =J_{ij}^{\top}r_{ij}, \\
    \tilde J_{ij}^{\top}\tilde J_{ij} & =J_{ij}^{\top}J_{ij}.
  \end{aligned}\label{eq:identities}
\end{equation}
These equalities preserve the squared residual norm, the gradient, and the Gauss-Newton matrix, respectively, with the same factor weight as in \cref{eq:objective}.

In order to achieve this, we will seek expansions of both point and planar residuals in the following form,
\begin{align}
  \underbrace{E}_{m \times l} = \underbrace{F}_{m \times k} \underbrace{B(X_i, X_j)}_{k \times l},
\end{align}
where $k \ll m$, $l$ is 1 for planar residuals and $3$ for point residuals, and $E$ is a row-wise matrix of residuals. Note that $B$ is state dependent and will need to be recomputed as pose estimates are updated. In contrast, $F$ only depends on the matched points and normals and is computed a single time. This allows for a QR decomposition, $F = QU$, to also be computed a single time, resulting in the compressed residual,
\begin{align*}
  \tilde{E} = U B(X_i, X_j).
\end{align*}

Next, we expound on the specific formulation of $F$ and $B$ for both point and planar residuals.

\subsection{Point-to-Point Residuals}
The world-frame residual in \cref{eq:point} can be rearranged as
\begin{equation}
  \begin{aligned}
    r^{\mathrm{pt}}_{ij,a} & = R_j p_{j,a} + t_j - R_i p_{i,a} - t_i                              \\
                           & = R_j p_{j, a} + t_j - R_i p_{i,a} - t_i + R_j p_{i,a} - R_j p_{i,a} \\
                           & = (R_j-R_i)p_{i,a} + R_j (p_{j,a}-p_{i,a}) +(t_j-t_i).
    \label{eq:point-rearranged}
  \end{aligned}
\end{equation}
The seven fixed coefficients of this expression form the row
\begin{equation}
  q_a=\begin{bmatrix}1&p_{i,a}^{\top}&(p_{j,a}-p_{i,a})^{\top}\end{bmatrix}
  \in\mathbb R^{1\times7}.
  \label{eq:pointrow}
\end{equation}
All matches share the same pose-dependent matrix
\begin{equation}
  B_{\mathrm{pt}}(X_i,X_j)=
  \begin{bmatrix}
    (t_j-t_i)^\top \\(R_j-R_i)^\top\\R_j^\top
  \end{bmatrix}\in\mathbb R^{7\times3},
  \label{eq:point-basis}
\end{equation}
so that $(r^{\mathrm{pt}}_{ij,a})^\top=q_aB_{\mathrm{pt}}$.
Stack the coefficient rows into $F_{\mathrm{pt}}$ and the transposed residuals into $E_{\mathrm{pt}}$:
\begin{equation}
  \begin{aligned}
    F_{\mathrm{pt}} & =\begin{bmatrix}q_1\\\vdots\\q_{m_{\mathrm{pt}}}\end{bmatrix}
    \in\mathbb R^{m_{\mathrm{pt}}\times7},                                                                                                                 \\
    E_{\mathrm{pt}} & =\begin{bmatrix}(r^{\mathrm{pt}}_{ij,1})^\top\\\vdots\\(r^{\mathrm{pt}}_{ij,m_{\mathrm{pt}}})^\top\end{bmatrix}
    =F_{\mathrm{pt}}B_{\mathrm{pt}}\in\mathbb R^{m_{\mathrm{pt}}\times3}.
  \end{aligned}\label{eq:point-factorization}
\end{equation}
Each row of $E_{\mathrm{pt}}$ contains the three coordinates of one match's residual.
Householder QR of $F_{\mathrm{pt}}$, padded with zero rows if $m_{\mathrm{pt}}<7$, yields a $7\times7$ upper-triangular matrix $U_{\mathrm{pt}}$ satisfying $U_{\mathrm{pt}}^\top U_{\mathrm{pt}}=F_{\mathrm{pt}}^\top F_{\mathrm{pt}}$.
We replace the original residual matrix with
\begin{equation}
  \tilde E_{\mathrm{pt}}=U_{\mathrm{pt}}B_{\mathrm{pt}}\in\mathbb R^{7\times3}.
  \label{eq:compressed-point}
\end{equation}
The compressed matrix has seven rows and three columns, hence 21 scalar residuals, regardless of the number of point matches.
We retain the point residuals in world coordinates because a pose-dependent rotation preserves their norm but can change their Gauss-Newton matrix.

\subsection{Point-to-Plane Residuals}
Define the relative rotation and translation by $R_{ij}=R_i^\top R_j$ and $t_{ij}=R_i^\top(t_j-t_i)$, and let $I_3$ be the $3\times3$ identity matrix.
The scalar residual in \cref{eq:plane} then becomes
\begin{equation}
  \begin{aligned}
    r^{\mathrm{pl}}_{ij,a} & = (R_i n_{i,a})^\top (R_j p_{j,a} +t_j- R_i p_{i,a}-t_i)                \\
                           & = n_{i,a}^\top(R_i^\top R_j p_{j,a} - p_{i,a} + R_i^\top (t_j - t_i) )  \\
                           & = n_{i,a}^\top (R_{ij}  p_{j,a} - p_{i,a} + t_{ij} + p_{j,a} - p_{j,a}) \\
                           & = n_{i,a}^\top(R_{ij}-I_3)p_{j,a}+n_{i,a}^\top t_{ij}
    +n_{i,a}^\top(p_{j,a}-p_{i,a}).
  \end{aligned}\label{eq:plane-rearranged}
\end{equation}
To collect its coefficients, let $\operatorname{vec}_r(A)$ stack the entries of a matrix $A$ row by row into a column vector.
For match $a$, define
\begin{equation}
  z_a=\begin{bmatrix}
    \operatorname{vec}_r(n_{i,a}p_{j,a}^\top) \\
    n_{i,a}                                   \\n_{i,a}^\top(p_{j,a}-p_{i,a})
  \end{bmatrix}^{\!\top}\in\mathbb R^{1\times13}.
  \label{eq:planerow}
\end{equation}
The corresponding pose-dependent vector is
\begin{equation}
  B_{\mathrm{pl}}(X_i,X_j)=
  \begin{bmatrix}\operatorname{vec}_r(R_{ij}-I_3)\\t_{ij}\\1\end{bmatrix}
  \in\mathbb R^{13},\quad
  r^{\mathrm{pl}}_{ij,a}=z_a B_{\mathrm{pl}}.
  \label{eq:planeeval}
\end{equation}
Its nine rotation entries are computed from $R_{ij}\in\mathrm{SO}(3)$ and do not introduce additional optimization variables.
Stacking all coefficient rows and scalar residuals gives
\begin{equation}
  \begin{aligned}
    F_{\mathrm{pl}} & =\begin{bmatrix}z_1\\\vdots \\ z_{m_{\mathrm{pl}}}\end{bmatrix}
    \in\mathbb R^{m_{\mathrm{pl}}\times13},                                                                                          \\
    E_{\mathrm{pl}} & =\begin{bmatrix}r^{\mathrm{pl}}_{ij,1}\\\vdots \\ r^{\mathrm{pl}}_{ij,m_{\mathrm{pl}}}\end{bmatrix}
    =F_{\mathrm{pl}} B_{\mathrm{pl}}\in\mathbb R^{m_{\mathrm{pl}}}.
  \end{aligned}\label{eq:factorization}
\end{equation}
Householder QR of $F_{\mathrm{pl}}$, padded with zero rows if $m_{\mathrm{pl}}<13$, gives $U_{\mathrm{pl}}\in\mathbb R^{13\times13}$ with $U_{\mathrm{pl}}^\top U_{\mathrm{pl}}=F_{\mathrm{pl}}^\top F_{\mathrm{pl}}$.
The compressed plane residual vector is
\begin{equation}
  \tilde E_{\mathrm{pl}}=U_{\mathrm{pl}} B_{\mathrm{pl}}\in\mathbb R^{13}.
  \label{eq:compressed-plane}
\end{equation}
We form $\tilde r_{ij}$ by listing the entries of $\tilde E_{\mathrm{pt}}$ row by row, followed by $\tilde E_{\mathrm{pl}}$.
It therefore contains $3\cdot7+13=34$ scalar residuals.%
An absent match type contributes no residuals.

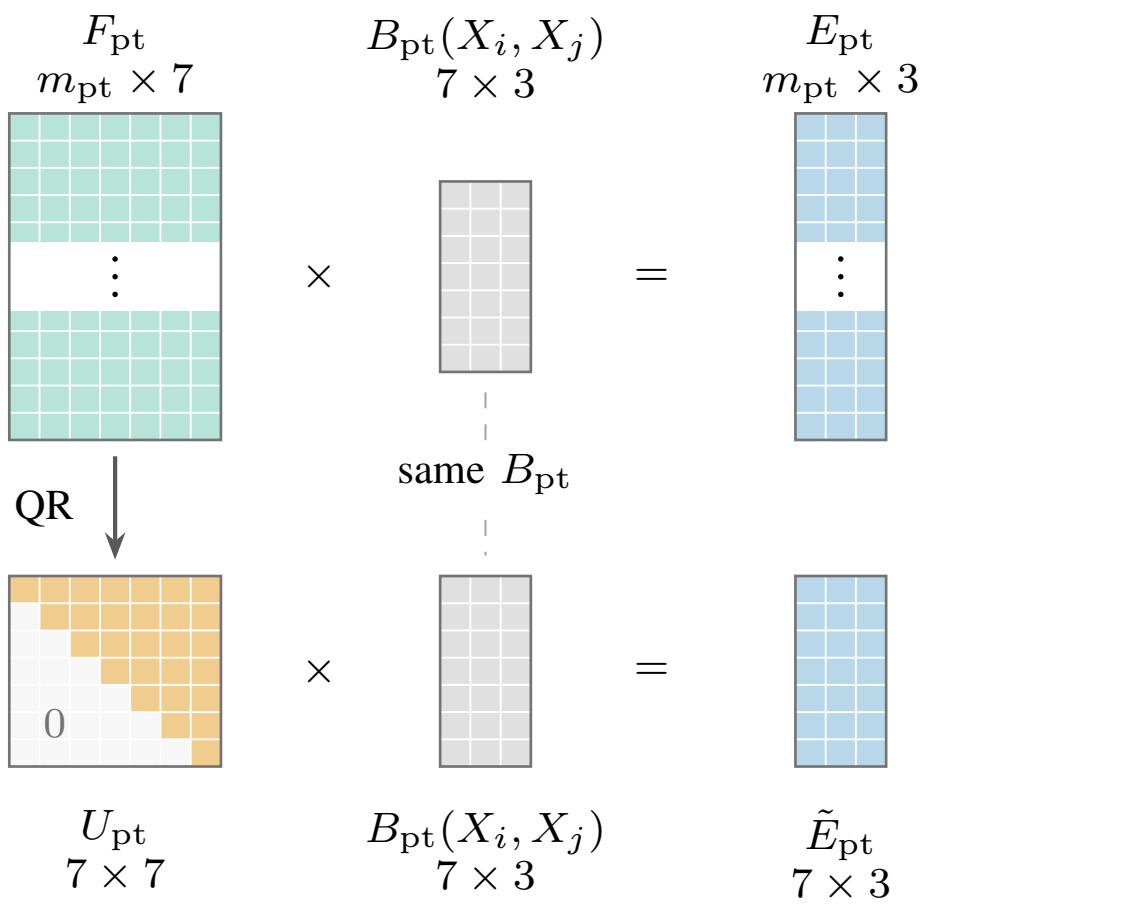
\begin{figure}[t]
  \centering
  \begin{tikzpicture}[x=1cm,y=.9cm,font=\footnotesize,
      gridline/.style={draw=white,line width=.35pt},
      outline/.style={draw=black!55,line width=.5pt},
      arr/.style={-{Stealth[length=1.8mm]},draw=black!65,line width=.8pt}]
    \fill[stMatch!28] (.15,2.5) rectangle (1.55,4.9);
    \fill[stExtract!28] (5.35,2.5) rectangle (5.95,4.9);
    \foreach \c in {1,...,6}
      \draw[gridline] ({.15+.2*\c},2.5) -- ({.15+.2*\c},4.9);
    \foreach \c in {1,2}
      \draw[gridline] ({5.35+.2*\c},2.5) -- ({5.35+.2*\c},4.9);
    \foreach \r in {1,...,11} {
      \draw[gridline] (.15,{2.5+.2*\r}) -- (1.55,{2.5+.2*\r});
      \draw[gridline] (5.35,{2.5+.2*\r}) -- (5.95,{2.5+.2*\r});
    }
    \fill[white] (.15,3.45) rectangle (1.55,3.95);
    \fill[white] (5.35,3.45) rectangle (5.95,3.95);
    \foreach \y in {3.57,3.70,3.83} {
      \fill[black] (.85,\y) circle[radius=.018cm];
      \fill[black] (5.65,\y) circle[radius=.018cm];
    }
    \draw[outline] (.15,2.5) rectangle (1.55,4.9);
    \draw[outline] (5.35,2.5) rectangle (5.95,4.9);

    \foreach \y in {3.0,.1} {
      \fill[black!12] (3,\y) rectangle (3.6,{\y+1.4});
      \foreach \c in {1,2}
        \draw[gridline] ({3+.2*\c},\y) -- ({3+.2*\c},{\y+1.4});
      \foreach \r in {1,...,6}
        \draw[gridline] (3,{\y+.2*\r}) -- (3.6,{\y+.2*\r});
      \draw[outline] (3,\y) rectangle (3.6,{\y+1.4});
    }

    \fill[black!3] (.15,.1) rectangle (1.55,1.5);
    \foreach \r in {0,...,6}
      \foreach \c in {\r,...,6}
        \fill[stSemi!45] ({.15+.2*\c},{1.3-.2*\r})
          rectangle ({.35+.2*\c},{1.5-.2*\r});
    \foreach \g in {1,...,6} {
      \draw[gridline] ({.15+.2*\g},.1) -- ({.15+.2*\g},1.5);
      \draw[gridline] (.15,{.1+.2*\g}) -- (1.55,{.1+.2*\g});
    }
    \node[text=black!55] at (.45,.42) {$0$};
    \draw[outline] (.15,.1) rectangle (1.55,1.5);

    \fill[stExtract!28] (5.35,.1) rectangle (5.95,1.5);
    \foreach \c in {1,2}
      \draw[gridline] ({5.35+.2*\c},.1) -- ({5.35+.2*\c},1.5);
    \foreach \r in {1,...,6}
      \draw[gridline] (5.35,{.1+.2*\r}) -- (5.95,{.1+.2*\r});
    \draw[outline] (5.35,.1) rectangle (5.95,1.5);

    \foreach \y in {3.7,.8} {
      \node at (2.2,\y) {$\times$};
      \node at (4.4,\y) {$=$};
    }
    \draw[arr] (.85,2.38) -- node[left=4pt] {QR} (.85,1.62);
    \draw[black!35,dashed] (3.3,2.85) -- (3.3,1.65);
    \node[fill=white,inner sep=2pt] at (3.3,2.25) {same $B_{\mathrm{pt}}$};

    \node[align=center] at (.85,5.3) {$F_{\mathrm{pt}}$\\$m_{\mathrm{pt}}\times7$};
    \node[align=center] at (3.3,5.3) {$B_{\mathrm{pt}}(X_i,X_j)$\\$7\times3$};
    \node[align=center] at (5.65,5.3) {$E_{\mathrm{pt}}$\\$m_{\mathrm{pt}}\times3$};
    \node[align=center,anchor=north] at (.85,-.08) {$U_{\mathrm{pt}}$\\$7\times7$};
    \node[align=center,anchor=north] at (3.3,-.08) {$B_{\mathrm{pt}}(X_i,X_j)$\\$7\times3$};
    \node[align=center,anchor=north] at (5.65,-.08) {$\tilde E_{\mathrm{pt}}$\\$7\times3$};
  \end{tikzpicture}
  \caption{Factor compression for point-to-point matches. QR replaces $F_{\mathrm{pt}}$ with the triangular summary $U_{\mathrm{pt}}$. Multiplication by the same pose matrix $B_{\mathrm{pt}}$ gives $3m_{\mathrm{pt}}$ original or 21 compressed residuals. Plane residuals similarly compress to thirteen scalars.}
  \label{fig:compression}
  \vspace{-1.5em}
\end{figure}

\subsection{Exactness and Linearization}
Both constructions use a coefficient matrix $F\in\mathbb R^{m\times k}$, with $(m,k)=(m_{\mathrm{pt}},7)$ or $(m_{\mathrm{pl}},13)$.
Let $M=\max(m,k)$ and let $\bar F\in\mathbb R^{M\times k}$ be $F$ padded with zero rows if necessary.
The QR factorization is
\begin{equation}
  \bar F=Q\begin{bmatrix}U\\0\end{bmatrix},\qquad
  Q\in\mathbb R^{M\times M},\quad Q^\top Q=I_M,
  \label{eq:qr}
\end{equation}
where $I_M$ is the identity matrix and $U\in\mathbb R^{k\times k}$ is upper triangular.
Consequently,
\begin{equation}
  F^\top F=\bar F^\top\bar F=U^\top U.
  \label{eq:gram-preservation}
\end{equation}
Both $F$ and $U$ depend only on the fixed matches, not on the poses.

We now show that this matrix identity preserves the quantities in \cref{eq:identities}.
Let $h(X_i,X_j)\in\mathbb R^k$ denote one column of $B_{\mathrm{pt}}$ for point matches, or $ B_{\mathrm{pl}}$ for plane matches.
The original and compressed residual vectors for this choice are $s=Fh$ and $\tilde s=Uh$.
To differentiate them, use the local increments defined in the overview as right-multiplicative perturbations
\begin{equation}
  X_b(\delta_b)=X_b\operatorname{Exp}(\delta_b),\quad
  \delta_b=(\omega_b^\top,v_b^\top)^\top,\quad b\in\{i,j\}.
  \label{eq:pose-update}
\end{equation}
Here $\omega_b,v_b\in\mathbb R^3$ are rotation and translation increments in pose $b$'s local frame, and $\operatorname{Exp}$ is the $\mathrm{SE}(3)$ exponential map.
Let $G=\partial h/\partial\delta\in\mathbb R^{k\times12}$, evaluated at zero increment.
Since $F$ and $U$ are fixed, the two residual Jacobians are $J_s=FG$ and $\tilde J_s=UG$.
Substituting \cref{eq:gram-preservation} gives
\begin{equation}
  \begin{aligned}
    \tilde s^\top\tilde s     & =h^\top U^\top Uh=h^\top F^\top Fh=s^\top s,     \\
    \tilde J_s^\top\tilde s   & =G^\top U^\top Uh=G^\top F^\top Fh=J_s^\top s,   \\
    \tilde J_s^\top\tilde J_s & =G^\top U^\top UG=G^\top F^\top FG=J_s^\top J_s.
  \end{aligned}\label{eq:channel-identities}
\end{equation}
Summing these contributions over the three point coordinates and the plane residuals proves \cref{eq:identities} for the complete scan-pair factor.
The common weight $1/\sigma_{ij}^2$ multiplies both sides equally.
The equalities hold at every pose pair, so the compressed residuals can be evaluated and relinearized after pose updates without returning to the individual matches.

\subsection{Use in FORM}
xFORM constructs a summary when its scan pair's matches change and reuses it while those matches remain fixed.
Pose updates change $B_{\mathrm{pt}}$ and $ B_{\mathrm{pl}}$ but leave $U_{\mathrm{pt}}$ and $U_{\mathrm{pl}}$ unchanged.
\Cref{sec:summary-stages} describes how these summaries are used in FORM's two optimization stages.

For $m\ge k$ matches with $k$ coefficients each, constructing a summary costs $O(mk^2)$ for QR.
Once constructed, evaluating its residuals and Jacobians costs $O(k^2)$ because the numbers of pose variables and residual coordinates per row are fixed.
Thus, the repeated evaluation cost no longer grows with the number of matches.

We compute Householder QR in double precision without forming $F^\top F$ or truncating numerical rank.
The expressions $p_{j,a}-p_{i,a}$ and $R_{ij}-I_3$ reduce cancellation near small residuals, with the latter evaluated as $R_i^\top(R_j-R_i)$.
The preservation guarantees assume fixed matches and FORM's common isotropic weight per scan-pair factor.
Pose-dependent robust weights or anisotropic noise would require a different construction.

\section{CPU and GPU Implementation}\label{sec:acceleration}

We combine factor compression with accelerated feature extraction and correspondence search to build two optimized variants of FORM.
xFORM-CPU runs entirely on the CPU, while xFORM-GPU also uses the GPU for feature processing, correspondence search, and summary construction.
Both retain FORM's feature-selection rules, registration objective, and optimization schedule.

\subsection{Feature Extraction}\label{sec:extraction}
FORM computes a curvature score for each point and selects planar candidates one scanline at a time.
Within each scanline, selecting a candidate excludes nearby points from subsequent selections.
xFORM-CPU processes different scanlines in parallel while preserving the selection order within each scanline.

xFORM-GPU performs curvature computation and feature selection on the GPU.
It identifies valid points, computes their curvature scores in parallel, and sorts points by curvature within each angular sector for planar-feature selection.
Planar and point-feature selection run in parallel across scanlines while preserving FORM's selection order within each scanline.

Normal estimation also runs on the GPU.
For each planar candidate, the GPU searches adjacent scanlines for their nearest valid points, gathers neighborhoods from those scanlines and the candidate's own scanline, and computes the normal by a covariance eigensolve.
The GPU returns the selected feature indices and estimated normals to the CPU.
The CPU retrieves the feature coordinates from the original scan, so intermediate curvature values and neighborhoods need not be transferred.

\subsection{Correspondence Search and Summary Construction}\label{sec:matching}
At each matching step, the new scan's features are matched to features of the same type in the local map.
xFORM-CPU retains FORM's correspondence search and constructs factor summaries for the updated scan pairs in parallel using Householder QR.
Summaries whose matches have not changed are reused.

xFORM-GPU performs both correspondence search and summary construction on the GPU.
FORM searches the same local map throughout the matching loop, using updated estimates of the new scan's pose.
xFORM-GPU uploads the retained map features, their poses, and the new scan's features once per scan and keeps the matching data on the device throughout the loop.
The GPU transforms the retained map features into world coordinates using their poses and constructs the local map used for search.
At each matching step, it transforms the query features into the world frame using the current pose and searches for their nearest map features in parallel.
The searches use FORM's distance threshold and tie-breaking rule.

To construct summaries without first downloading the matches, the GPU groups accepted matches by the source scan of the matched map feature.
Each group defines a scan pair.
The GPU then forms the coefficient rows in \cref{eq:pointrow,eq:planerow} using the matched points and normals in their original scan frames.
It computes QR for blocks of these rows in parallel, then stacks and factors the resulting triangular matrices until one summary remains for each match type in each scan pair.
These QR computations use double precision.

The GPU returns the summaries and match counts to the CPU for optimization, avoiding a transfer of individual correspondences at every matching step.
The final correspondences are transferred after the matching loop for FORM's subsequent constraint and map updates.

\subsection{Pose Optimization Using Factor Summaries}\label{sec:optimizer}\label{sec:summary-stages}
FORM uses the GTSAM library to linearize registration factors and assemble the linear system for the pose update.
xFORM computes each scan pair's contribution directly from its factor summaries.
For pair $(i,j)$, let $\tilde r_{ij}\in\mathbb R^{d_{ij}}$, with $d_{ij}\le34$, be its compressed residual vector.
Let $\tilde J_{ij}\in\mathbb R^{d_{ij}\times12}$ be its Jacobian with respect to the two local pose increments defined in \cref{eq:pose-update}.
The pair's Gauss--Newton matrix and negative gradient are
\begin{equation}
    H^{(ij)}=\frac{\tilde J_{ij}^\top\tilde J_{ij}}{\sigma_{ij}^2},\qquad
    g^{(ij)}=-\frac{\tilde J_{ij}^\top\tilde r_{ij}}{\sigma_{ij}^2}.
    \label{eq:pair-block}
\end{equation}
xFORM computes these contributions in parallel across scan pairs.
It adds each $H^{(ij)}$ to the global matrix at the rows and columns of poses $i$ and $j$, and each $g^{(ij)}$ to the corresponding entries of the global vector.
This avoids constructing intermediate GTSAM linearized registration factors at each LM iteration.
The remaining factors are unchanged.
The LM loop retains FORM's damping, acceptance, pose-update, and stopping rules.

FORM's two optimization stages (\cref{sec:form}) differ in which registration factors are relinearized.
During semi-linearized optimization, factors involving the newest scan are evaluated and relinearized from their summaries at each LM iteration.
Factors between older scans retain a quadratic approximation formed at the window's pose estimates after processing the preceding scan.
After matching finishes, full optimization evaluates and relinearizes every registration factor from its summary.
In both variants, the iterative LM loop evaluates summaries and assembles and solves the pose system on the CPU for the evaluated windows of at most \MaxActivePoses{} poses, below the configured GPU threshold of 75 poses.

\section{Experimental Evaluation}\label{sec:experiments}

\subsection{Setup}\label{sec:setup}
We evaluate end-to-end latency and numerical agreement with FORM in a main comparison, then examine the contributions of individual changes through separate ablation runs.
All experiments use an AMD Ryzen 9 5900X with 12 worker threads and an NVIDIA GeForce RTX 5090, with release builds using GCC 13.4, Eigen 3.4, CUDA 12.8, and GTSAM 4.2.
We use Hilti 2022~\cite{zhangHiltiOxfordDatasetMillimeterAccurate2023} (Hilti, 32 scanlines), Multi-Campus~\cite{nguyenMCDDiverseLargeScale2024} (MCD, 64/128 scanlines), and Newer College 2021~\cite{zhangMultiCameraLiDARInertial2022} (N21, 128 scanlines).
Inputs are prepared with evalio 0.6.1~\cite{potokarComprehensiveEvaluationLiDAR2025} without deskewing, using range limits of 0.5--120\,m, 0.1--120\,m, and 0.1--50\,m, respectively.
All configurations share the same inputs and FORM settings: 3 point and 50 planar features per sector, feature spacing 5, and 10 recent scans.

Timing covers the complete estimator, including GPU transfers and synchronization, but excludes input loading, output serialization, and the first 20 scans.
Correspondence-search time includes map preparation and GPU summary construction; optimization time includes semi-linearized and full optimization, excluding marginalization.
All evaluated LM linear systems are solved on the CPU.

\subsection{Main Comparison}\label{sec:main}
We compare FORM, xFORM-CPU, and xFORM-GPU on five complete sequences per dataset.
Each configuration runs twice on every dataset, with execution order reversed in the second repeat (\BenchmarkRunCount{} runs total).
Dataset mean and 95th-percentile latencies (p95) average the corresponding per-run statistics equally across sequences and repeats.
Speedup is the ratio of FORM's mean latency to that of each xFORM variant.

\subsubsection{End-to-End Latency}
\begin{table}[t]
\centering
\caption{Full-sequence latency (ms per scan), using five sequences per dataset. Mean and p95 weight sequences equally, with two runs per configuration on every dataset, excluding 20 initial scans per run. Speedup is the ratio of mean latencies.}
\label{tab:latency}
\scriptsize
\setlength{\tabcolsep}{2.6pt}
\begin{tabular}{lrrrrrrrr}
\toprule
 & \multicolumn{2}{c}{FORM} & \multicolumn{3}{c}{xFORM-CPU} & \multicolumn{3}{c}{xFORM-GPU} \\
Dataset & mean & p95 & mean & p95 & speedup & mean & p95 & speedup \\
\midrule
N21 & 112.20 & 172.61 & 90.75 & 139.67 & 1.24$\times$ & 30.85 & 51.47 & 3.64$\times$ \\
Hilti & 56.86 & 135.56 & 43.59 & 102.28 & 1.30$\times$ & 25.32 & 61.50 & 2.25$\times$ \\
MCD & 71.27 & 105.00 & 59.79 & 88.93 & 1.19$\times$ & 23.46 & 39.20 & 3.04$\times$ \\
\bottomrule
\end{tabular}
\end{table}

On average across all fifteen sequences, xFORM-CPU and xFORM-GPU achieve speedups of \OverallCpuSpeedup{}$\times$ and \OverallGpuSpeedup{}$\times$ over FORM, respectively (\cref{tab:latency}).
Across individual sequences, xFORM-CPU achieves \CpuSpeedupRange{}$\times$ speedups and xFORM-GPU achieves \GpuSpeedupRange{}$\times$.

\subsubsection{Accuracy and Numerical Agreement}\label{sec:agreement}
\begin{table}[t]
\centering
\caption{Measured accuracy and numerical agreement on five full sequences per dataset. RTE (m) is root-mean-square translation error over 1/30 m segments, averaged across sequences and two repeats. FORM and both xFORM variants agree at the shown precision. Maxima compare xFORM poses against FORM at identical timestamps, without alignment.}
\label{tab:agreement}
\scriptsize
\setlength{\tabcolsep}{2.6pt}
\begin{tabular}{lrrrr}
\toprule
Dataset & RTE 1 m & RTE 30 m & Max $\Delta$pos (m) & Max $\Delta$rot (rad) \\
\midrule
N21 & 0.0613 & 0.4756 & $4.3\times10^{-10}$ & $8.4\times10^{-12}$ \\
Hilti & 0.1182 & 1.0265 & $1.9\times10^{-10}$ & $1.9\times10^{-11}$ \\
MCD & 0.0871 & 0.5478 & $1.2\times10^{-8}$ & $3.4\times10^{-11}$ \\
\bottomrule
\end{tabular}
\vspace{-1.5em}
\end{table}

Across all scans in the main comparison, xFORM-CPU and xFORM-GPU agree with FORM to within $\AgreementPosition$\,m in position and $\AgreementRotation$\,rad in rotation (\cref{tab:agreement}).
These maxima compare poses at identical timestamps against FORM from the same repeat, without trajectory alignment.
All per-scan feature, match, active-pose, scan-pair, rematch, and LM-iteration counts agree across the \TraceComparisonCount{} paired comparisons, as do the matched ground-truth sets.
The table reports our measured translational relative trajectory errors (RTE) over 1\,m and 30\,m segments, using the root-mean-square definition of FORM~\cite{potokarFORMFixedLagOdometry2025}.
We average each sequence's RMS error over its two repeats, then weight the five sequences equally within each dataset.
FORM and both xFORM variants agree at the displayed precision.
Individual-sequence accuracy and ground-truth coverage are provided in the supplementary tables.

\subsection{Ablation Studies}\label{sec:ablation}
\subsubsection{Cumulative CPU and GPU Acceleration}
\begin{table}[t]
\centering
\caption{Cumulative ablation on three individual complete sequences: Stairs (Newer College 2021), Basement 2 (Hilti 2022), and NTU Day 01 (Multi-Campus). Columns report sequence results, not dataset averages. Each row adds to the preceding configuration. Values are mean latency (ms per scan) from one run per configuration, excluding the first 20 scans. The first three rows use only the CPU.}
\label{tab:ablation}
\footnotesize
\setlength{\tabcolsep}{3pt}
\begin{tabular}{>{\raggedright\arraybackslash}p{.53\columnwidth}rrr}
\toprule
Configuration & Stairs & \shortstack{Basement\\2} & \shortstack{NTU\\Day 01} \\
\midrule
FORM & 134.78 & 47.46 & 92.58 \\
\;+ optimization using factor summaries & 110.78 & 36.10 & 78.78 \\
\;+ parallel planar selection \textbf{(xFORM-CPU)} & \textbf{107.71} & \textbf{33.76} & \textbf{75.90} \\
\;+ GPU matching and summary construction & 43.32 & 23.97 & 49.39 \\
\;+ GPU curvature and neighbor search & 36.35 & 19.22 & 45.13 \\
\;+ GPU normal computation & 35.22 & 19.04 & 43.96 \\
\;+ GPU feature selection and matching preparation \textbf{(xFORM-GPU)} & \textbf{32.55} & \textbf{17.26} & \textbf{26.20} \\
\bottomrule
\end{tabular}
\end{table}

We measure the cumulative contributions of optimization, feature extraction, and correspondence-search changes on three complete sequences: N21 Stairs, Hilti \texttt{basement\_2}, and MCD \texttt{ntu\_day\_01} (\cref{tab:ablation}).
Each of the seven configurations runs once per sequence, giving 21 runs collected separately from the main comparison under the same settings and timing protocol.
Each row adds to the preceding configuration, and each column reports one sequence.
The first addition combines factor compression with the optimizer implementation in \cref{sec:optimizer}.

Parallel planar-feature selection reduces extraction from \AblationStairsFormExtract{} to \AblationStairsCpuExtract{}\,ms, but correspondence search still accounts for \AblationStairsCpuMatch{} of xFORM-CPU's \AblationStairsCpuMean{}\,ms total.
GPU matching and summary construction provide the largest reduction among the GPU additions on this sequence, lowering total latency to \AblationStairsGpuMatchingMean{}\,ms; GPU feature processing and the remaining matching preparation reduce it further to \AblationStairsGpuMean{}\,ms.

\subsubsection{Factor Compression Acceleration}
\begin{table}[t]
\centering
\caption{Factor compression acceleration on the complete N21 Stairs sequence. Mean time in ms per scan; speedup is relative to FORM. The compressed variant retains original CPU feature extraction and correspondence search. Optimization includes CPU summary construction. These are the same two runs as the first two Stairs entries in \cref{tab:ablation}.}
\label{tab:optimizer-ablation}
\footnotesize
\setlength{\tabcolsep}{3pt}
\begin{tabular}{>{\raggedright\arraybackslash}p{.32\columnwidth}rrr}
\toprule
Mean time/frame & \shortstack{Original\\FORM} & \shortstack{Factor\\compression} & Speedup \\
\midrule
Semi-linearized opt. & 19.54\,ms & 10.98\,ms & 1.78$\times$ \\
Full-nonlinear opt. & 13.00\,ms & 0.91\,ms & 14.21$\times$ \\
Combined opt. & 32.54\,ms & 11.89\,ms & 2.74$\times$ \\
Entire pipeline & 134.78\,ms & 110.78\,ms & 1.22$\times$ \\
\bottomrule
\end{tabular}
\vspace{-1.5em}
\end{table}

To examine factor compression acceleration, we compare original FORM with the factor-summary optimizer in \cref{sec:optimizer}, retaining original CPU feature extraction and correspondence search in both configurations (\cref{tab:optimizer-ablation}).
We reuse the first two configurations on Stairs from the cumulative ablation: one run each over all 1,190 scans, with 1,170 timed scans after warmup.
Semi-linearized optimization decreases from \AblationOptimizerSemiForm{} to \AblationOptimizerSemiCompressed{}\,ms per frame (\AblationOptimizerSemiSpeedup{}$\times$), while full-nonlinear optimization decreases from \AblationOptimizerFullForm{} to \AblationOptimizerFullCompressed{}\,ms (\AblationOptimizerFullSpeedup{}$\times$).
Including CPU summary construction, their combined time decreases from \AblationOptimizerCombinedForm{} to \AblationOptimizerCombinedCompressed{}\,ms (\AblationOptimizerCombinedSpeedup{}$\times$).
The larger proportional gain in full-nonlinear optimization reflects repeated relinearization of all factors, whereas semi-linearized optimization already reuses older linearizations.
With original feature extraction and correspondence search retained, entire-pipeline latency decreases from \AblationOptimizerPipelineForm{} to \AblationOptimizerPipelineCompressed{}\,ms per frame (\AblationOptimizerPipelineSpeedup{}$\times$).

\section{Conclusion}\label{sec:conclusion}
\vspace{-0.25em}
xFORM accelerates FORM through exact factor compression and parallel CPU and GPU processing.
For fixed matches, factor compression represents each scan pair with a $7\times7$ point summary and a $13\times13$ plane summary, preserving the factor's cost and Gauss--Newton system at any poses in exact arithmetic.
Evaluating and linearizing these summaries costs the same regardless of the match count.
xFORM-CPU combines parallel evaluation and linearization of factor summaries with parallel feature selection to achieve a \CpuSpeedupRange{}$\times$ speedup over FORM.
xFORM-GPU further accelerates feature extraction, correspondence search, and summary construction, achieving an overall speedup of \GpuSpeedupRange{}$\times$.
Both variants closely reproduce FORM's trajectories.

\par
\bibliographystyle{ieeetr}
\footnotesize
\bibliography{ref}
\end{document}